# Artificial Intelligence without Restriction Surpassing Human Intelligence with Probability One: Theoretical Insight into Secrets of the Brain with AI Twins of the Brain

Guang-Bin Huang[1,2,7*], M. Brandon Westover[3§], Eng-King Tan[4,6†], Haibo Wang[5†], Dongshun Cui[7], Wei-Ying Ma[8], Tiantong Wang[9], Qi He[5], Haikun Wei[1,2], Ning Wang[7], Qiyuan Tian[10], Kwok-Yan Lam[9,11], Xin Yao[12,13], Tien Yin Wong[14,15*]

[1] School of Automation, Southeast University, Nanjing, China
[2] Key Laboratory of Measurement and Control of Complex Systems of Engineering, Ministry of Education, Nanjing, China
[3] Beth Israel Deaconess Medical Center, Harvard Medical School, Boston, USA
[4] Department of Neurology, National Neuroscience Institute, Singapore
[5] Research Centre of Big Data and Artificial Intelligence for Medicine, First Affiliated Hospital of Sun Yat-Sen University, Guangzhou, China
[6] Duke-NUS Medical School, National University of Singapore, Singapore
[7] Mind PointEye, Singapore
[8] Institute for AI Industry Research, Tsinghua University, Beijing, China
[9] College of Computing and Data Science, Nanyang Technological University, Singapore
[10] School of Biomedical Engineering, Tsinghua University, Beijing, China
[11] Singapore AI Safety Institute, Nanyang Technological University, Singapore
[12] School of Data Science, Lingnan University, Hong Kong SAR, China
[13] School of Computer Science, University of Birmingham, UK
[14] Singapore Eye Research Institute, Singapore National Eye Centre, Singapore
[15] Tsinghua Medicine, Tsinghua University, Beijing, China

*Corresponding authors. G.-B. Huang: gbhuang@ieee.org, gbhuang@seu.edu.cn; T. Y. Wong: wongtienyin@tsinghua.edu.cn
†These authors contributed equally to this work
§Professor M. B. Westover is a co-founder, scientific advisor, consultant to, and has personal equity interest in Beacon Biosignals, USA
#This paper has been accepted for publication in Neurocomputing (26 November 2024)

**Abstract:** Artificial Intelligence (AI) has apparently become one of the most important techniques discovered by humans in history while the human brain is widely recognized as one of the most complex systems in the universe. One fundamental critical question which would affect human sustainability remains open: Will artificial intelligence (AI) evolve to surpass human intelligence in the future? This paper shows that in theory new AI twins with fresh cellular level of AI techniques for neuroscience could approximate the brain and its functioning systems (e.g. perception and cognition functions) with any expected small error and AI without restrictions could surpass human intelligence with probability one in the end. This paper indirectly proves the validity of the conjecture made by Frank Rosenblatt 70 years ago about the potential capabilities of AI, especially in the realm of artificial neural networks. This paper also gives the answer to the two widely discussed fundamental questions: 1) whether AI could have potentials of discovering new principles in nature; 2) whether error backpropagation (BP) algorithm commonly and efficiently used in tuning parameters in AI applications is also adopted in the brain. Intelligence is just one of fortuitous but sophisticated creations of the nature which has not been fully discovered. Like mathematics and physics, with no restrictions artificial intelligence would lead to a new subject with its self-contained systems and principles. We anticipate that this paper opens new doors for 1) AI twins and other AI techniques to be used in cellular level of efficient neuroscience dynamic analysis, functioning analysis of the brain and brain illness solutions; 2) new worldwide collaborative scheme for interdisciplinary teams concurrently working on and modelling different types of neurons and synapses and different level of functioning subsystems of the brain with AI techniques; 3) development of low energy of AI techniques with the aid of fundamental neuroscience properties; and 4) new controllable, explainable and safe AI techniques with reasoning capabilities of discovering principles in nature.

I. Introduction

Artificial Intelligence (AI) has experienced three important development phases: following its warmup stage of development in the 1950s to 1980s (Stage I) and research driven period from the 1980s to 2010 (Stage II), AI has transitioned into a golden industrial data driven era since 2010 (Stage III). There is no doubt that after 70 years of development, it has been a new normal that AI frequently surpasses human performance in numerous applications. Thus, the extensively discussed and widely argued question turns out to become critical: Would AI without appropriate restriction and governance surpass human intelligence in the end?

It is widely recognized that human brains may be one of the most complicated systems in the universe. The human body receives signals from its environment mainly through the five types of sensors (human sense organs: eyes, ears, noses, tongues, and skins). The corresponding five human senses are sight, hearing, smell, taste, and touch. These signals are sent to the brain. The human brain, a sophisticated biological system, processes and interprets sensory signals for perception including vision, hearing and temperature, and manages cognitive recognition through learning, memory, reasoning, thought and emotion regulation to control all regulatory functions of the human body.

The human brain is composed of hundreds of different types of biological neurons (biological cells) through large number of interconnected biological connections (synapses):
1) The number of biological neurons in the human brain is huge but finite, e.g., about 86 billion neurons for average adults with about 16 billion neurons in human's cerebral cortex[1]. Hundred types of neurons have been discovered in human brains[2].
2) Synapses are the junctions through which neurons connect and communicate with each other in the brain. The number of synapses in the brain is finite as well. Each neuron has a finite number of synaptic connections, e.g., from a few to hundreds of thousands of synaptic connections, with itself (through autapses), neighboring neurons, or neurons crossing various regions of the brain. Autapses are synapses between a neuron's axon and its own dendrites[3]. The human brain contains about hundreds of trillion synapses[4]. Different from chemical synapses, volume transmission and ehaptic coupling may occur in signaling transmissions among neurons. Electrical synapses (gap junctions) may allow electrical signals to pass directly from one neuron to another[5].
3) The number of neurons and synapses are crucible parameters for human intelligence[6].

There would forever have no direct answer to the critical question when two levels of research complexity are intertwined:
1) *Macro system level*: The brain is acknowledged as a complex system cooperated by billions of neurons and trillions of synapses with feedback as a whole.
2) *Micro system level*: Neurons and synapses are expected to have explainable representations through mathematical modeling / formula[7,8] and molecular behaviors[9], and research mainly focuses on discovering such mathematical representations and molecular behaviors for each type of these large number of neurons and synapses from the perspective of neuroscience dynamics[8,10].

This paper shows that the breakthrough and concrete answers could be given to this critical question by delving into the indivisible functional level of the brain's fundamental components, such as neurons and synapses, and recognizing their unique properties along with AI techniques. Typically, instead of focusing on studying neuroscience using traditional neuron dynamics techniques, each of the finite number of neurons and synapses of the brain could be replaced and represented by corresponding AI components sequentially. In theory, regions and functioning subsystems (e.g., visual systems, olfactory systems, hearing systems, reasoning systems, etc) of the brain, which may have infinite number of unknown functions, could be universally approximated by corresponding AI twins with any expected small error (see the Brain-AI-Representation Theorem in Section IV).

## II. Challenges of classical mathematical modelling and neuro dynamics approaches

The structure of the human brain is intricate and complicated. Neuroscientists have sought to divide the brain into smaller pieces[11] in order to understand how the brain works as a whole. It is difficult and challenging to fully understand the sophisticated learning mechanism, infinite number of unknown functions and philosophy of the human brain based on its pieces and regions with feedback of signals. Pieces and regions of the brain are composed of neural circuits which may feedback signals in complex ways.

Spiking neurons and neuron dynamics are often used to describe neurons' characters. Both the spiking neuron models and neuron dynamics methods aim to provide mathematical descriptions of the conduction of electrical signals in neurons, e.g., the role of the biophysical and geometrical characteristics of neurons on the conduction of electrical activity, including spatial morphology or the membrane voltage dynamics[12]. One of most typical neuron models is the Nobel Prize awarded Hodgkin–Huxley model (H&H model)[13,14,15,16] which expresses the relationship between the flow of ionic currents across the neuronal cell membrane and the membrane voltage of the cell with a set of nonlinear differential equations. For example, the voltage-current relationship of a typical neuron can be given as

$$I = C_m \frac{dV}{dt} + I_i$$

where $I$ is the total current density through the membrane, $I_i$ is the ionic current density, $C_m$ is the membrane capacity per unit area, and $t$ is time[14]. Multiple parameters may be estimated or measured for an accurate model of a biological neuron and it would be difficult to have accurate models for all types of biological neurons in the end.

With approximately 86 billion neurons and hundreds of trillions of synapses exhibiting these complex dynamics, it is nearly impossible to easily analyze such a system without detailed understanding and modellings of neurons, synapses and their subtypes. In general, to focus on the details of spiking neurons and their dynamic behaviors could make the analysis of the brain become complicated. Classical mathematical modellings need to consider the unique features and characteristics of neurons and synapses, basic neuroanatomical and functional divisions.

The electrical and chemical properties / transmissions are hugely different. The synapse is the holy grail of neuroscience research. The neurochemical transmission through the release, diffusion, receptor binding of neurotransmitters plays vital roles in brain functions. The regulation and speed of termination and regeneration involving the reuptake of proteins / transmitters, enzymatic degradation in the cleft and diffusion are key determinants of many neurological activities and functions. Communications between neurons and nonneural cells is also essential for axonal conduction, synaptic transmission, and information processing[17].

Mathematical modeling usually needs to consider the type of neuronal subtypes in the different regions and the metabolic activities also affect the biological vulnerability of the various neuronal subtypes. In addition, the resultant excitatory and inhibitory effects are determined by the type of neurotransmitter, whether it is excitatory (such as glutamate), inhibitory (such as GABA), or mixed (such as dopamine), and are also affected by the type of receptor bindings.

Neuroplasticity, which is the ability of the neurons to adapt structurally and functionally in response to intrinsic or extrinsic stimuli, is an interesting phenomenon. The reorganizing abilities especially after external (injuries, toxins) or intrinsic insults (microenvironmental stress) form the basis of many of the interventional measures.

The nervous system is more extensive as a whole. There is the peripheral nervous system, where both its afferent and efferent inputs can affect brain functions and modulate neuroplasticity. For example, if a peripheral limb is amputated, there will be changes in the relevant brain regions. There is also the autonomic nervous system which innervates the major organs such as the heart and the secretory glands.

### III. Divide-and-Conquer approach: General unified mathematical representations of neurons and synapses (as basic learning elements) of the brain

Different levels of systems in the brain are nonlinear and dynamical. Advanced mathematical modelling including almost endless possibilities and functions makes concise analysis of the brain impossible. The resultant accumulated error between the integrated mathematical models and the brain functions would not be controllable due to inaccurate estimation of each model for each type of neurons. To understand the brain and to discover its secrets seems impossible if AI research focuses solely on the high-level (e.g., brain, regions, pieces levels) systems of the brain. Surprisingly, breakthroughs could happen when the unique properties of fundamental neurons and synapses are considered together with cellular level of AI techniques. AI opens a door to study the secrets of the brain due to its ability of approximating nonlinear functions of fundamental neurons and synapses as well as the systems and subsystems built on top of them.

*Foundation of Brain's recursive mechanisms: Four-distinctive-properties of neurons and synapses*

Intriguingly, from the perspective of a divide-and-conquer approach, when we delve down to the indivisible function level of the brain in which pieces and regions cannot be divided further, the human brain turns out to be a remarkably beautiful and elegant recursively structured system, and a groundbreaking approach to analyzing the learning structure of the brain could be developed. Especially, our research work finds that all the following four-distinctive-properties are necessarily put together to consider.

*Four Fundamental Properties of Building Blocks of the Brain*:
1) *Two fundamental components*: The brain's fundamental signaling and communication functions are primarily driven by two key components: biological neurons and biological synapses.
2) *Unidirectional signal transmission*: Both neurons and synapses transfer signal to other neurons in one direction[18] (Fig. 1).
3) *Alternative connections in sequence*: Biological neurons and biological synapses are alternately connected one after another in sequence.
4) *"All-or-None Law"*: Neurons follow "all-or-none law." If a neuron responds, it must respond completely[19].

In a neural circuit formed by a group of neurons which are interconnected by synapses, one of the neurons may send a signal back to some of its initiating neurons. However, this may follow the unidirectional transmission property that signals transfer from one neuron to another in such a linear sequence[20]. Such a unidirectional signal transmission property and "all-or-none law" play a critical role in the learning system of the brain. These four key properties distinguish brains from other types of electrical circuits and systems, and essentially build links and fill up the gap between biological learning and AI. Recapitulating these key properties of the biological neurons and synapses together with the synergy of the renowned learning capabilities of AI[21,22,23,24], one could analyze the learning capabilities of the brain recursively with the combinations of divide-and-conquer and bottom-up approaches. Similar analysis can be linearly extended to volume transmission, ehaptic coupling and electrical synapses (gap junctions) as well as direction connections among neurons.

Every single synapse is made up of a presynaptic terminal and a postsynaptic terminal. The presynaptic terminal at the end of an axon of a neuron converts the electrical signal (the action potential of a neuron) into a chemical signal (neurotransmitter release) and the neurotransmitter diffusing across the synaptic cleft finally binds to receptors in the postsynaptic terminal. A single neuron may receive thousands of input signals through its dendrite trees. After integrating all the received signals (information) the postsynaptic neuron further decides whether it fires or not based on its own internal action potential mechanism[25,26].

Action potentials depend on special types of voltage-gated ion channels of neurons' membranes[20]. In most cases, the relationship between membrane potential and channel state is probabilistic and is expected to have a time delay.

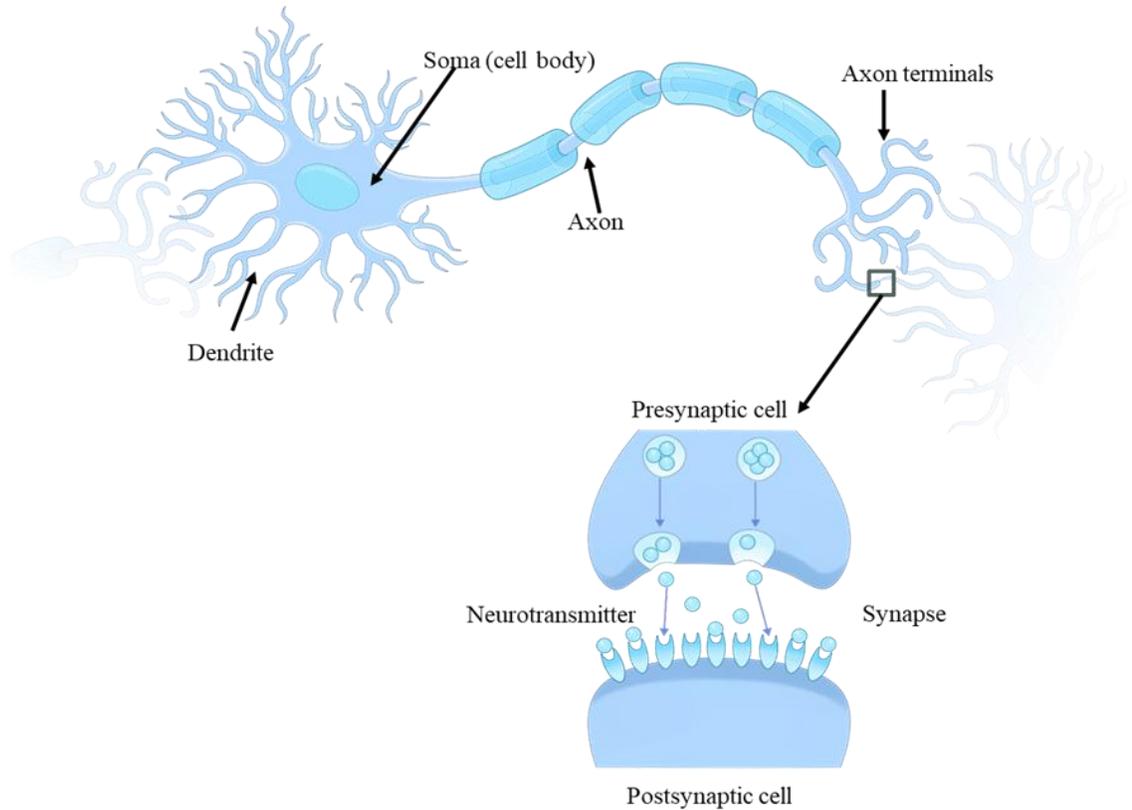
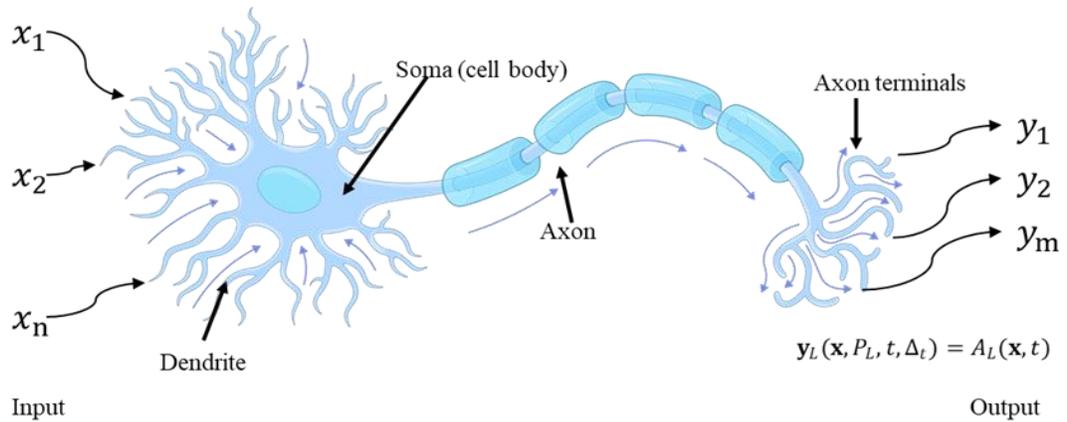

**Fig. 1. Unidirectional signal transmission with fundamental components (neurons and synapses) alternatively connected in sequence.** (**A**) Biological neurons and synapses in the brain with unidirectional signal transmissions from one neuron to another and within a synapse, both neurons and synapses provide unidirectional signal transmissions. (**B**) Unidirectional signal transmission from one neuron to another, the probability $P_L$ and time delay $\Delta_t$ in the relationship between membrane potential and channel state are considered as piecewise continuous functions.

*General unified mathematical representations of fundamental components: neurons and synapses*

Different from studies on neuron characterstics[27,28,29,30,31,32] and neuron dynamics as well as those AI models focusing on divisional functions of pieces /regions / systems in the brain[33,34], this paper provides a novel AI twin approach which applies AI models to represent single neurons and synapses without knowing details of conduction of electrical signals in neurons and synapses and related mathematical modelling and molecular behaviors in single neurons and synapses. Hundreds of types of single neurons and their subtypes which are considered encapsulated fundamental building components of the brain could be represented by hundreds of types of corresponding AI models, in which AI works as new materials and components to implement and represent biological neurons. So do synapses.

It would be an endless effort to figure out the exact mathematical models and molecular behaviors of huge number of different types of neurons, synapses and their subtypes. However, no matter what types of essential functionality of the spiking representation and neurons dynamics mechanisms a neuron may have, general speaking the functionality of neurons and synapses may be considered piecewise continuous[8,10,35] due to their (lower frequencies) biological responses which are reasonably smooth to received signals. In principle, there may have three fundamental piecewise continuous functions within neurons:
1) Although the information encoding of neurons is still unknown, the information encoding mechanism and related spiking probabilities could be considered a function which is piecewise continuous.
2) Time delay variables follow some piecewise continuous mechanisms.
3) Signal transmission functions of neurons are piecewise continuous if considered together with probability function.

In essence, whether a single neuron can be excited into firing depends on the integrated input signals received through its dendrite trees. Neurons adhere to the all-or-none law. In other words, if a single neuron is excited, it will always give a maximal response and produce an electrical impulse of a single amplitude. It gives a maximal response or none at all. In this sense, the input-output relationship functions between the input signals of neuron dendrites and the output signals of axon terminals, which is a composite function of action potential function and axon mechanism, is a nonlinear piecewise function of all the excitatory and inhibitory signals received by the neuron.

Suppose that a biological neuron (indexed by *L*) has input signal $\mathbf{x}_L$ in its dendrite trees and the output signal of its presynaptic terminal is $\mathbf{y}_L(\mathbf{x}) = A_L(\mathbf{x})$, where $A_L(\mathbf{x})$ shows the signal transferring functions (including the probability and time delay in such relationship between membrane potential and channel state, see Appendix A) between the input signals of neurons' dendrites and the output signals of axon terminals (Fig. 1B). In essence, the signal transferring functions $A_L(\mathbf{x})$ between the input signals of dendrites and the output signals of axon terminals are naturally (piecewise) continuous as well due to biological reaction properties of neurons.
This provides an alternative solution and representation of biological neurons which is different from the conventional spiking representation and neurons dynamics approaches[8].

Like the action potentials of neurons, the neurotransmission relationship of $P$-th synapse between the presynaptic terminal at the end of an axon of a neuron and the postsynaptic terminal of the next connected neuron is $S_P(\mathbf{x})$, which is (piecewise) continuous as well[35].

In principle, the human brain could be considered as a function of a set of two fundamental elements (signal transferring relationships of single neuron $A_L$ between its dendrite trees and its axon terminal, neurotransmission relationship within corresponding synapses $S_P$), that is $(A_L, S_P)$, where $L = 1, \cdots, 86$ billion (or around 86 billion). In this sense, despite the intricate nature of brain structures, brains turn out to be such a simple, efficient, beautiful, and elegant system:
   1) From the *architectural point of view*, brains are composed of two fundamental elements: neurons and synapses.
   2) From the *theoretical point of view*, the brain system is a composite combination of two fundamental (piecewise) continuous functions: signal transferring relationship of a single neuron, which governs the signal transferring between the input signals received by its dendrites and the output signals sent from its axon terminals $A_L$ and neurotransmission relationship within corresponding synapses $S_P$.

### IV. Bottom-Up approach: General learning system of the brain and its regions / subsystems represented with AI Twins

Ensembles of neurons amalgamate into physiological regions and yield functional properties and states in the brain[36]. The human brain system (including learning, memory, reasoning, thought, feeling, emotion, vision, hearing, etc) is constructed from (piecewise) continuous combination of finite number of the signal transferring relationships of neurons $A_L(\mathbf{x})$ and the neurotransmission relationships within their corresponding synapses $S_P(\mathbf{x})$. That is, with bottom-up approach the human brain could be viewed as a straightforward and elegant structure constructed like building blocks from layers and groups of neurons and synapses, each layer and group built upon other layers and groups.

In contrast to the commonly recognized and applied spiking representation and neuron dynamics mechanisms for biological neurons, one could leverage the learning and modelling capabilities of AI techniques (instead of mathematical modelling methods) to approximate neurons without requiring detailed knowledge of their biological characteristics. Since both fundamental elements (components) are (piecewise) continuous and they are connected alternatively sequentially in order, with bottom-up recursive approach the human brain system is (piecewise) continuous as well (See Theorem 4 in Appendix C).

*AI approximating single neuron and single synapse*

It is known that if the hidden node activation function of single-hidden layer feedforward networks (SLFNs) is bounded nonconstant continuous, then any target continuous function $f$ can be approximated by such SLFN with any expected small error[21,22,24]. That is, given any small positive value $\delta > 0$, there exists SLFN with $f_L$ with appropriate number $L$ of hidden nodes and properly adjusted hidden node parameters such that $|f - f_L| < \delta$.

Without loss of generality, single-hidden layer feedforward networks (SLFN) are used in the basic theoretical analysis in this paper and all the analysis can be linearly replaced with more efficient (and more complicated) AI models whenever necessary. Thus, given any signal transferring relationships of single neuron $A_L(\mathbf{x})$ and the neurotransmission relationships within their corresponding synapses $S_P(\mathbf{x})$, which are piecewise continuous, there exists an AI model (e.g., artificial feedforward neural networks) universally approximating them. In other words, the two fundamental computational elements (e.g., neurons and synapses) which form the human brain learning systems could be represented by an efficient artificial neural network with any expected small error. These artificial neural networks could effectively learn and represent a wide spectrum of relationships among neurons and the neurotransmission processes within corresponding synapses. Since both fundamental signal transferring relationships of single neuron $A_L(\mathbf{x})$ and the neurotransmission relationships within their corresponding synapses $S_P(\mathbf{x})$ are (piecewise) continuous, they could be represented by AI with networks of artificial neurons. Thus, we have

1) Given any signal transferring relationship of a single neuron between its dendrite trees and the presynaptic terminal of its axon $A_L(\mathbf{x})$, there exists an artificial neural network which could universally approximate $A_L(\mathbf{x})$ with any expected small error (See Theorem 1 in Appendix B).
2) Given any neurotransmission relationship within corresponding synapses $S_P(\mathbf{x})$, there exists an artificial neural network which could universally approximate $S_P(\mathbf{x})$ with any expected small error (see Theorem 2 in Appendix B).

These essentially suggest that in theory for any biological neuron and synapse within the human brain, there exists an efficient AI model capable of universally approximating them, e.g., the signal transferring relationship of any single neuron between its dendrite trees and the presynaptic terminal of its axon $A_L(\mathbf{x})$ and the neurotransmission relationship within corresponding synapses $S_P(\mathbf{x})$.

In practical AI applications, hybrid approaches that incorporate artificial neurons along with other mathematical formulations or smart materials are commonplace. These efficient artificial neural networks are implemented with various architectures, including but not limited to fully connected feedforward networks, convolutional neural networks[37,38], transformers[39], extreme learning machines (ELMs) [22,40,41], etc. In theory, out of these AI techniques, learning algorithms of extreme learning machines (ELMs) have universal approximation capabilities[22,40,41], and logically speaking, could be used for the universal approximation of biological neurons and synapses.

*AI approximating learning capabilities of the brain and its sequentially linked regions / subsystems*

Neurons adhere to the all-or-none law that if a neuron responds it must respond completely. The functions of synapses could be considered as all-or-none smoothness although the outputs of hundreds of trillions of synapses may not have a single amplitude and vary significantly. The meaning of all-or-none smoothness (see Definition 3 in Appendix C) extends a little more from all-or-none law in neurons and does not require the output of function has uniform value (amplitude) due to some facts:
1) Nonlinear functions of synapses may not have uniform values.
2) Functions and systems in the level of pieces / regions of the brain are usually not uniform.

3) Each period of firings and inhibitions of single biological neurons is limited. Thus, it may not have infinite number of oscillations / spiking in neurons and signal transmission in synapses for any reasonable duration, and the nonlinear functions of single neurons and synapses are piecewise continuous.
4) Neurons may communicate through changes in membrane potentials (graded potentials), which are close to "analog" signals rather than "discrete" spikes[41,43,44,45].

In this context, the psychological all-or-none law observed in neurons represents a specific instance of the broader mathematical all-or-none smoothness function:
1) If a biological neuron responds after receiving signals from its dendrite trees, it must respond completely[19].
2) Physiologically, the all-or-none law shows the principle that if a single neuron is excited, its axon will always generate an electrical impulse of uniform amplitude. This impulse height remains constant regardless of the stimulus intensity or duration. Neuron axons either transmit a maximal response across the synapse to the next neuron entirely or not at all.
3) The all-or-none smoothness functions include not only the psychological all-or-none law observed in the neurons of the brain but also the piecewise continuous functions in synapses. It is valid for the majority of commonly utilized practical functions in real-world applications (mathematical basis function, basic mathematical formula, smart materials, etc.). Uniform amplitude is not required in these practical functional implementations of AI for single neurons.

A general fact (see Theorem 3 in Appendix C) is that when all-or-none smoothness piecewise continuous functions are combined with another all-or-none smoothness piecewise continuous function, the resultant composite function retains the all-or-none smoothness piecewise continuous property. This obviously includes the cases when biological neurons and synapses are sequentially alternatively connected one by one in order and when such combination of all-or-none smoothness piecewise continuous functions is applied recursively in such sequences. The brain and any of its constituent regions sequentially constructed by such sequentially linked unidirectional biological neurons and synapses preserve the all-or-none smoothness piecewise continuous property (Fig. 2C) (see Theorem 4 in Appendix C). That is, from Theorem 3 in Appendix C we have

**General Theorem of All-or-None Smoothness Preservation for Brain** (All-or-None Smoothness Preservation Theorem)
Suppose that all the functions $f, f_1, \cdots, f_k$ are all-or-none smoothness piecewise continuous, then $f(f_1, f_2, \cdots, f_k)$ are all-or-none smoothness piecewise continuous. And thus, the human brain and any of its constituent regions constructed by sequentially linked neurons and synapses in the brain are All-or-None smoothness piecewise continuous.

Function $f_k$ could be generated from such sequences with very few to large number of neurons and synapses. Such sequences may contain loops (Fig. 2C and Fig. 2D) although neurons and synapses are unidirectional. In theory, the human brain and any of its regions and subsystems (*sequentially*) constructed by neurons and synapses and the corresponding functions could be represented and universally approximated by artificial neural networks (one of typical AI techniques) with any expected small error when each of its neurons and synapses is sequentially replaced and represented by appropriate AI components (Theorem 5 in Appendix C) (Fig. 2). Thus, from Theorem 5 in Appendix C we have

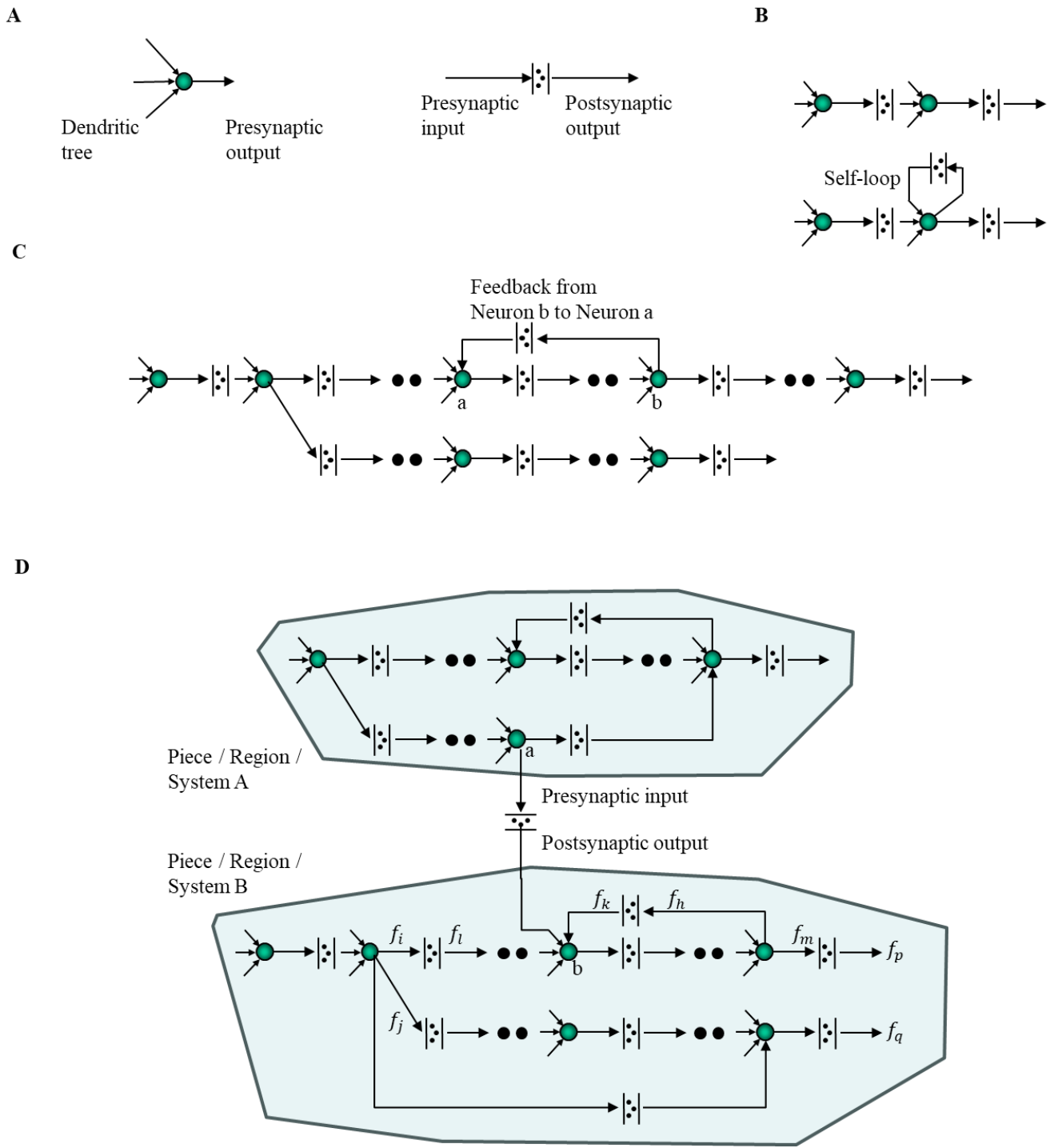

**Fig. 2. Brain Representation by AI Circuits (BrainAIC) – a new AI twin model of the brain: brain with AI structural sequential representation, e.g, Brain and its neuron and synapse circuit systems with AI as fundamental components.** (**A**) Two fundamental components of the

brain: Neurons and synapses, which are represented by AI. Three dots "." show signal transmission directions in synapses. Each subtype of fundamental neurons and synapses could be represented by one type of AI model. (**B**) Unidirectional signal transmission within the two fundamental components of the brain. Self-loop of neurons remains unidirectional. (**C**) Unidirectional neurons and synapses alternatively linked in sequence. (**D**) BrainAIC: New graph representation with neurons as vertices and synapses as edges. Regions / subsystems of the brain represented by AI circuits with AI components which sequentially replace unidirectional neurons and synapses, each of these AI components in BrainAIC could be implemented by smart materials, AI chips etc. Each component in electrical circuits is considered and studied individually with its own properties, even though its inputs and outputs depend on the inputs and outputs of other connected components. Similarly, biological neurons and synapses are considered encapsulated fundamental building components of the brain could be represented by corresponding AI models. Similar analysis could apply to different types of connections among neurons (e.g., chemical synapses, volume transmission, ehaptic coupling, electrical synapses, etc).

**General Theorem of Brain Representation by AI Twins** (Brain - AI Representation Theorem)
Given any function $f$ of a (sub-)system and region with sequentially linked neurons and synapses of the brain and any small positive value $\delta > 0$, there exists an AI twin $f_{AI}$ such that $|f - f_{AI}| < \delta$, when each of neurons and synapses in such a (sub-)system and region is sequentially replaced and represented by appropriate AI components. In other words, for the brain and any of its regions and subsystems which are constructed by *sequentially linked* neurons and synapses, there exist corresponding AI twins which could represent and universally approximate their functions with any small error.

*AI Twins of the brain*

It is difficult to define the meaning of human intelligence and to figure out its infinite number of functions. Intuitively speaking, there may have infinite number of unknown perception (vision, hearing, temperature, touching, etc) and cognition (thinking, memory, learning, problem-solving, language, and decision-making, etc) functions $f$ in the brain and there would have no way to understand and represent all "unknowns" if based on conventional mathematical and neuron dynamic methodologies. It would be impractical to exhaustively compare artificial intelligence (AI) and human intelligence function by function if AI models are based on brain functioning regions and subsystems rather than being built bottom-up as AI twins from physical components (e.g., neurons, synapses, etc).

In an electrical circuit, each component is considered and studied individually with its own properties, even though its inputs and outputs depend on the inputs and outputs of other connected components. One could replace each of its components without knowing the entire functionality of the circuits while remaining the functions of the electrical circuits unchanged. Similarly, the brain is a physical system which is different from abstract mathematical systems, different functions $f$ in the brain result from corresponding sequences of *physically* linked biological neurons and synapses (including electrical synapses, volume transmission and ehaptic coupling, etc). In theory the general theorem of brain representation is valid for all such functions $f$ which are built by physical architectures with sequentially linked neurons and synapses in the brain. Thus, in theory bottom-up approach based AI twins proposed in this paper could have similar capabilities of the brain. In this

Page **12** of **33**

context, to address the critical question whether AI would surpass human intelligence, it may not be necessary to identify and understand all the different types of functions generated in the brain when biological neurons and synapses would be sequentially represented by AI models one by one.

**V.      Complementary capabilities of AI and the Brain**

With appropriate governance, AI could provide complementary positive support to the ongoing technological revolution, which will ultimately benefit humans.

*AI as positive techniques with appropriate governance*

In addition to tasks commonly undertaken by both AI and humans, some tasks could be handled by AI alone and some principles in nature could be discovered by AI and humans with the aid of AI as well (Fig. 3).

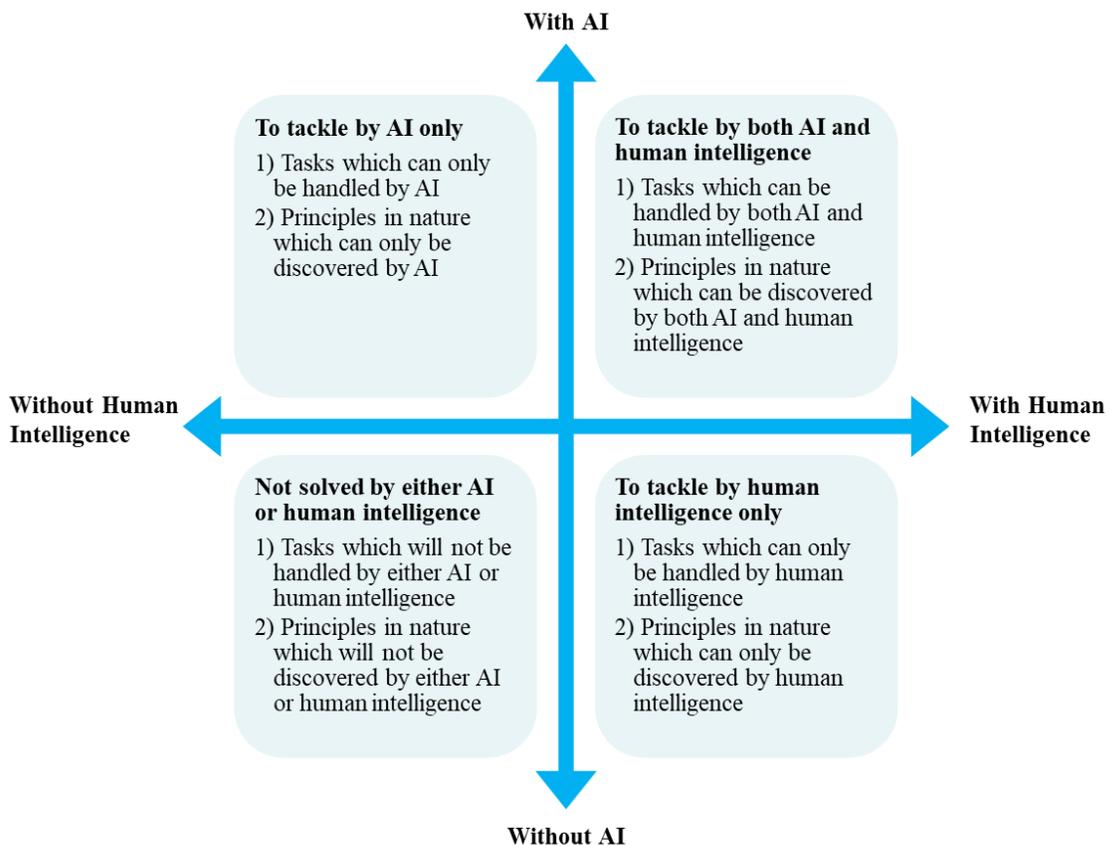

**Fig. 3. Four quadrants of the nature.** From intelligence perspective, there are four categories of discovery capabilities of nature when AI is under appropriate restriction and governance. Intelligence is composed of biological intelligence, artificial intelligence (AI) and the intelligence further self-evolved from artificial intelligence.

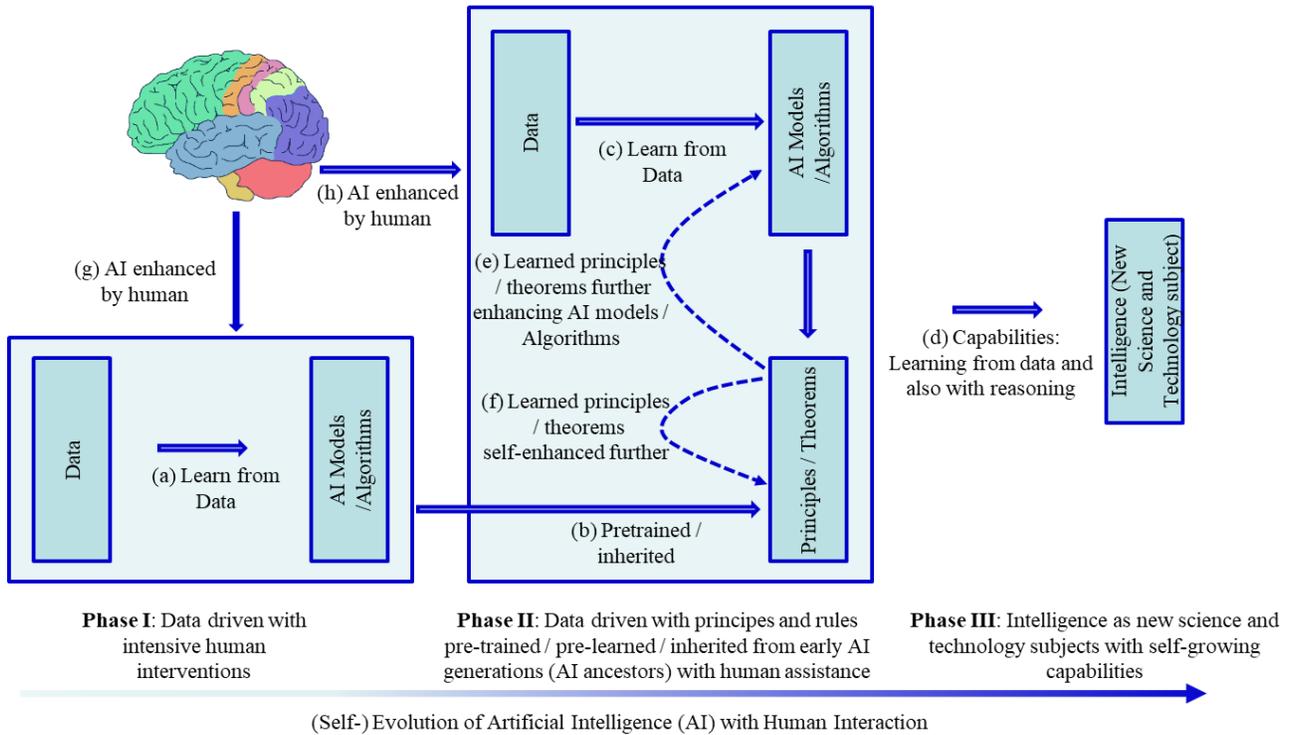

**Fig. 4. AI's potential of discovering new principles and theorems.** Since in theory AI twin as a specific AI model could universally approximate the brain and its functioning systems with any expected small error, in general AI would have the potential of discovering principles and theorems existing in nature. Phase I: AI learns from data (a) (c) with human interventions. Phase II: AI models with pretrained principles / theorems / facts (e) (f) could be inherited to further enhance their performance and generate advanced AI models adapted to new environments. Phase III: AI is paving the way for a new domain of scientific and technological subject in Intelligence (d).

In general, positive aspects of AI also include providing alternative solutions and potentials to discovering new nature principles (Fig. 4).
1) *AI's perception and cognition capabilities*. In theory AI twins as one of specific AI models could approximate the brain and its functioning systems with any expected error. If AI develops without restriction, the ultimate situation is that the brain could be represented and approximated by AI twins and other AI systems (Brain-AI-Representation Theorem in Section IV), and AI would have capabilities similar to brains such as perception and cognition capabilities (including learning, memory, reasoning, thought, emotion, etc). Thus, with the aid of understanding functioning systems of the brain, new controllable and explainable AI techniques have appropriate and positive perception and cognition capabilities could be developed to accordingly.
2) *AI's lower energy and small data implementation*. On the other hand, in the early stage of AI, functions and modellings which represent the essence of nature behind the data are learned by AI techniques, and thus large amount of data and extensive computing power with large amount of energy power are required. Since AI has such ultimate brain representation capabilities, when more and more essences and core principles of natures are learned by AI, AI may have less dependance on large amount of data and extensive computing resources (computing power and energy) in future evolutions. AI could then further discover (new) principles and theorems recursively through its self-learning and adaptation to new data and required data could be as small as requested by the brain (Fig. 4). Something could be discovered by the human, something could be discovered by AI, something could be discovered by both, and something may not be discovered forever (Fig. 3).

Driven by Watt's rotary steam engine, the industrial revolution transferred productions from manual methods to machines, leading to a substantial increase in productivity and contributing significantly to the growth of the human population. Similarly, AI may herald a new era of revolution: the Intelligent Revolution, which accelerates the expansion of the digital economy (Fig. 5). This intelligent economy may be poised to surpass the scale of the traditional economy by multiple folds.

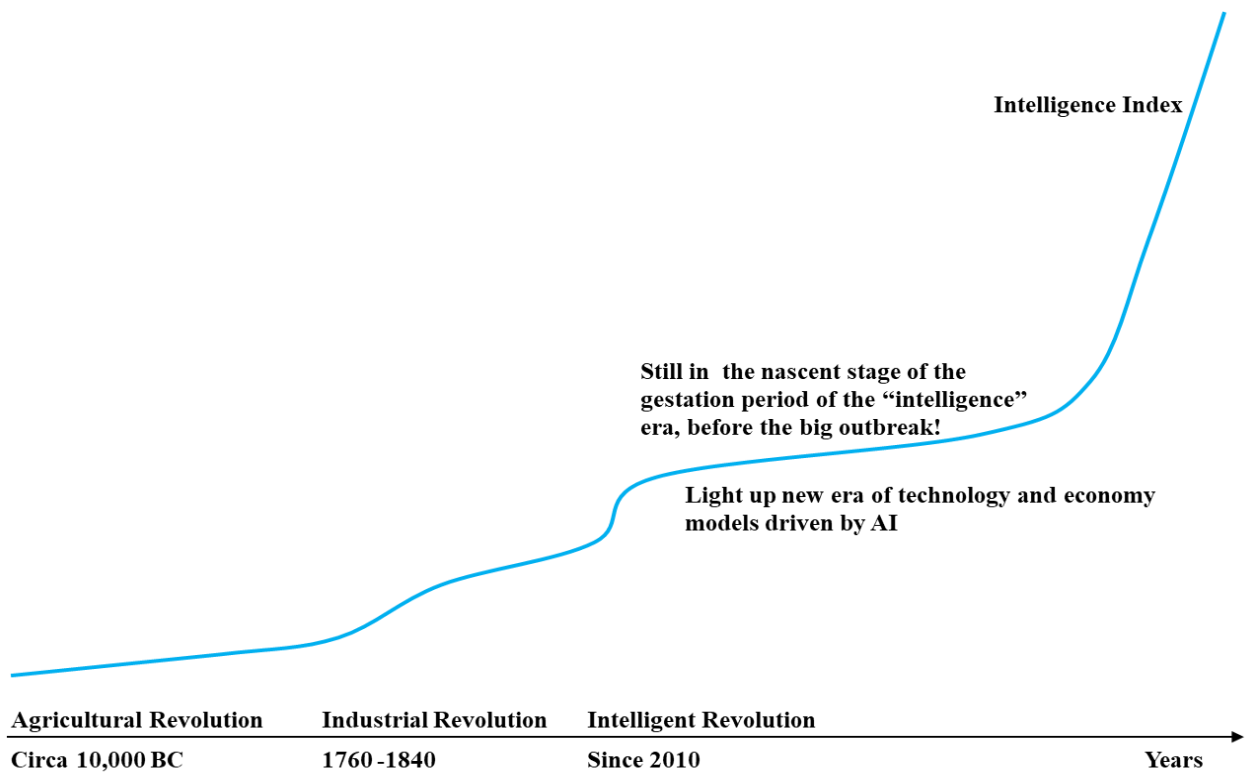

**Fig. 5. AI Era.** Analogous to agricultural revolution and industrial revolution, Intelligent revolution initiated by AI would play important roles in the history.

*AI surpassing human intelligence if with no restriction*

Theoretically for any given single neuron and synapse, there exist AI models capable of universally approximating them. Through bottom-up solutions, it is theoretically feasible for combinations of such AI models (AI twins) to universally approximate the corresponding capabilities of the brain and its sequentially linked regions / subsystems expected small errors. Moreover, theoretically, there may exist more efficient AI models capable of approximating any such functioning systems in the brain, rather than relying on stacks of AI models. Thus, if there were no proper restrictions and governances (see Appendix E), in the end AI could surpass human brain intelligence with probability one together with AI's exponential growth driven by the following key factors (to list a few), leading to serious concerns of existential risk of AI systems and redlines that need to be safeguarded by AI safety measures:

1) *Algorithms*: More and more advanced and efficient AI network architectures and algorithms will be employed in practical applications. Artificial neural networks or components of the AI networks / solutions could be replaced with more efficient or significantly simpler alternatives, e.g., various types of efficient network architectures, signal processing techniques and mathematical solutions (Fig. 6). Advanced AI algorithms may self-evolve further.

2) *Data Sources*: AI sensors may possess significantly larger sensing capabilities compared to human sensory organs (eyes, ears, noses, tongues, and skins). Consequently, AI could have access to a much broader source of data than humans.
   - AI could "hear" something which humans cannot "hear", e.g., AI could perceive some spectrum of sounds that are imperceptible to humans.
   - AI could "see" something which humans cannot "see", e.g., AI could detect some spectrum of visual information that is invisible to humans.
   - AI could "smell" something which humans cannot "smell", e.g., AI could detect some type of odors that are undetectable to humans.
   - AI could "taste" something which humans cannot "taste", e.g., AI could detect tastes that the human tongue cannot perceive.
   - AI could "feel" something which humans cannot "feel", e.g., AI could perceive some types of sensations that are beyond human capability to feel.

3) *Computing Power*: AI solutions could be implemented in computers, servers, devices, sensors, chips, etc. In theory, an infinite number of computing chips, sensors, devices, and servers could be connected and integrated in the world (through laboratories, organizations, regions, countries, etc), which makes the total computing power of AI beyond any single brain. Furthermore, rapidly emerging computing techniques (e.g., quantum computing) could implement algorithms and solve complex problems hundreds of millions of times fast than classical computers, shortening the required computing times from tens of thousands of years to minutes.

4) *Smart Materials*: AI could also be implemented and supported by wide types of smart materials including but not limited to neuromorphic, photonics, memristors, phase change materials, nano materials, etc. Integration of smart materials into AI chips could lead to significant improvements in performance, energy efficiency, and functionality, enabling the development of more powerful and versatile AI systems than ever.

5) *AI Agents*: AI could be implemented and deployed almost everywhere, including but not limited to AI chips, sensors, devices, robots, processes, systems, clouds, etc. Compared to human population, an enormous number of AI agents would be deployed in the future.

6) *Knowledge Exchange and Inheritance*: Human generation typically spans around 20 years or more, with individuals inheriting capabilities from their ancestors. Humans share and exchange knowledge in different means including various social activities. In contrast, AI has the capability of exchanging knowledge autonomously and could be updated continuously every moment, unlike the 20-year generational cycle observed in humans.

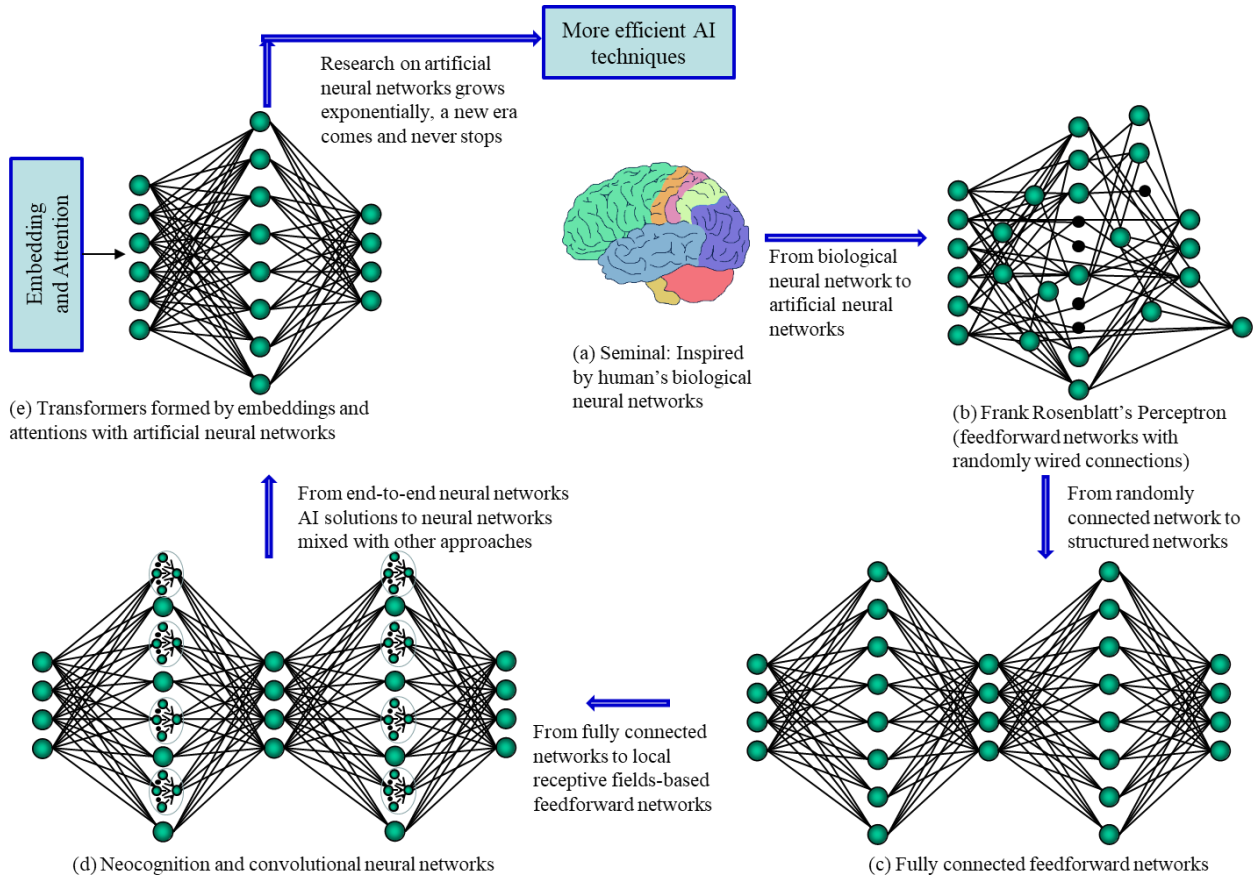

**Fig. 6. Evolution of artificial neural networks.** Some of typical artificial neural networks based AI inspired by biological neural networks (brains). This paper manages to theoretically prove the validity of the conjecture and expectation on the cognition capability of artificial neural networks which was made by Rosenblatt in a press conference around 65 years ago[46].

## VI. Discussions

It has been over 70 years since Rosenblatt introduced Perceptron[47], one of the first artificial neural networks inspired by biological principles. Over the past 70 years, out of many types of artificial neural networks, fully connected networks, convolutional neural networks, extreme learning machines and transformers have become popular. Based on Rosenblatt's statements in a 1958 press conference, it was reported by The New York Times the perceptron to be "the embryo of an electronic computer that will be able to walk, talk, see, write, reproduce itself and be conscious of its existence."[46] Since AI could approximate the brain and its functioning systems with any expected small error, in this sense this paper may have also theoretically proved the validity of Rosenblatt's conjecture and expectation.

In theory AI twins could be used to analayze the brain and its functioning systems (including learning, memory, reasoning, thought, feeling, emotion, vision, hearing, etc) with any expected small errors. In addition, AI further benefits from some key factors, e.g., algorithms, data sources, smart materials, AI agents, and knowledge exchanges for AI, etc. Thus, AI could surpass human intelligence if AI would develop without appropriate governance and restriction. It would be worth investigating how different types of cells / synapses and functioning systems could be modelled by AI techniques in the future and such research would request collective efforts from experts in multidisciplinary domains.

*AI as new alternative techniques disclosing secret of the brain*

It would forever have no way for us to understand the secret of the brain if the research is only driven by conventional mathematical modelling or neuroscience dynamics approaches. New mechanism phenomena which may have never been thought relevant to neurons and synapse functions could be found in different regions and subsystems of the brain[48]. It may be unachievable to find or manually derive mathematical models which could precisely represent all the functions of billions of neurons and trillions of synapses in the brain, let alone the subtypes of functions and systems of different pieces and regions of the brain. Any small inaccurate mathematical model representation in elementary levels could lead to unforeseen big differences in higher levels of systems in the brain. Sticking to traditional manpower intensive approaches would neither reveal the secret of the brain nor provide a direct answer to the fundamental question whether AI with no restrictions surpasses human intelligence in the end. However, the groundbreaking to the fundamental question could finally occur when human invented AI techniques as alternative new modelling technologies are used to help analyze and model the brain, with bottom-up approach recursively from modelling each type of neurons and synapses of the human brain to its functions and subsystems. This paper reveals the secret of the brain in four aspects:
1) From the fundamental element point of view, each type / subtype of neurons and each type / subtype of synapses could be represented and universally approximated by corresponding cellular level of AI models.
2) From the system architecture point of view, if ethic issues have been addressed, theoretically, AI could model brain architectures using bottom-up approaches due to the sequential links of finite number of unidirectional neurons and synapses in the brain, and the corresponding AI based neuron circuits could be built in the level of pieces, regions and systems. The brain could be considered as a complicated graph with neurons as vertices and synapses as edges, in which functions resulting in intelligence could be considered readouts of corresponding paths / links of sequentially connected neurons and synapses and proposed Brain Representation by AI Circuits (BrainAIC) could be considered as an AI mirror of such a brain graph with corresponding AI components modelling and representing biological neurons and synapses. Similar AI representations could be proposed for volume transmission, ehaptic coupling as well as electrical synapses (gap junctions)[5].
3) From the learning capabilities point of view, in theory the bottom-up built AI model could represent and universally approximate the functions of pieces, regions and systems in the brain with any expected small error (see the Brain-AI-Representation Theorem in Section IV).
4) From the technical methodology point of view, the brain system is recursive and elegant. Alternative sequential connections of unidirectional biological neurons and synapses play critical roles and form the foundation of the recursive approaches proposed in this paper for the analysis of the brain's structures and capabilities using AI techniques. This analysis is also valid to the direct connections between neurons and neurons.

In this sense, artificial intelligence could reach human level capabilities and eventually surpass human intelligence, and human intelligence would become a small subset of Intelligence if AI would grow exponentially without appropriate governance (Fig. 7 and Fig. 8). Thus, AI would self-grow and evolve after a certain stage, leading to a new domain and subject of Intelligence with its self-contained systems and principles.

*AI insights to brain energy saving and nature selection*

In addition to the fundamental problem question whether AI with no restrictions surpasses human intelligence in the end, this paper manages to give answer to two more challenging questions:
1) *Whether error backpropagation (BP) algorithm[49] is used in the brain*. BP algorithm is one of most popular mathematical methods used to efficiently update parameters in artificial neural networks especially deep learning. One primary difference between artificial neural networks and human brains lies in how error signals are delivered[50]. This paper perhaps manages to give the answer to an open challenging question whether BP algorithm is used in the brain. Mathematically derived BP algorithms commonly used in tuning parameters in AI applications require bidirectional information transmissions, e.g., forwarding acquired information and backwarding error signal over the same connections (connections among artificial neurons / computational nodes) [49] (see Appendix D). Although feedback connections are ubiquitous in the brain, the information transmission in both biological neurons and biological synapses of the human brain may be unidirectional. Thus, unlike artificial computational nodes and connections in artificial neural networks, it may be difficult to backward errors over biological neurons and biological synapses one by one although backward may appear in synapses or axons[50,51,52,53]. BP algorithms which require bidirectional information transmissions may not be feasible over sequentially linked unidirectional biological components (neurons and synapses) in the brain. BP algorithm involves a substantial number of calculations and iterations in order to converge, resulting in high energy consumptions. For instance, the average adult human brain consumes approximately 20 watts per day, whereas the frontier supercomputer requires 21 megawatts per day - equivalent to the power supply of about one million humans although performing quite different tasks[54]. Consequently, one plausible explanation for the absence of BP algorithm in the human brain is the necessity of energy efficiency. Such a high energy demand would not be viable for survival. This raises the intriguing question: how does the human brain adjust its "parameters" (including its "structures") to adapt to new information or correct errors? The human brain evolves over generations[8,9,55], retains advantageous biological traits or "parameters" to enhance survival and reproduction. Each generation fine-tunes its brain which is pre-trained in its ancestors. In essence, the human brain has evolved and adjusted its 'parameters,' including its structures, over millions of years through the process of natural selection and adaptation. This evolutionary process spans many generations, in stark contrast to the relatively short-term adjustments and rapid learning that occur within AI systems. One of the objectives of the AI training process assisted with BP algorithms is to learn knowledges which the human brain has achieved in its the extensive evolutionary journey into a much shorter period. Evolution mechanism of human brain intelligences shed lights on energy efficient implementation of artificial intelligence.
2) *Why spikes are used in the brain*. This paper manages to provide an intriguing interpretation of spikes for neurons, which may be attributed to natural selection[8]. Since all neurons and synapses with unidirectional transmission functionalities are sequentially alternatively connected one by one in order, this implies that neuron spiking encoding mechanism is possibly like frequency modulation. It is known that frequency modulation is more efficient than amplitude modulation. Such frequency modulation of encoding mechanism of biological neurons has several advantages:
    - It allows signal to be transferred over long distance with large numbers of neurons and synapses in the brain. There is no amplitude attenuation issue in such encoding mechanism.
    - It is more noise resistant and energy efficient than amplitude modulations.
    - Synapses can effectively pass signals among neurons.

*AI as new subject domain of the nature*

AI techniques grow exponentially mainly due to learning capabilities of artificial neural networks. Such networks may be composed of different types of artificial neurons, mathematical basis (e.g., Fourier series)[22,40,41,56], and different types of mathematical approaches (e.g., attention etc.)[39]. Analogous to circuits systems, contemporary neural network-based AI is giving rise to Networked Mathematics, a novel mathematical discipline focusing on interconnected networks (Fig. 2 and Fig. 6) of basis functions / computational nodes (including artificial neurons, smart materials, biological neurons and synapses in the brain) and AI agents / modules /systems, alongside a new field of Intelligence.

*AI as new alternative technique for neuroscience and brain illness*

This paper opens new doors for AI techniques to be used in cellular level of efficient neuroscience dynamic analysis and brain illness solutions.
1) A huge number of activity patterns in the brain build the foundation for adaptive behaviors. Understanding the complex neural dynamics mechanism with conventional approaches by which the human brain's hundred billion neurons and hundred trillion synapses manage to produce cortical configurations in a flexible manner continues to be a core challenge in neuroscience[4]. Divide-and-conquer and bottom-up based AI approaches are used simultaneously to potentially provide efficient solutions for neuroscience and brain analysis. AI twins and other AI technologies could be used to model large types of cell level of neurons and synapses and are then further used to learn the normal physiologic functions (including facilitatory and inhibitory signals) of each well-defined anatomic brain region (such as frontal lobe, midbrain, cerebellum, etc.) with integrated signals from the whole brain under varied external stimuli. Such AI techniques together with smart materials and AI chips techniques lead to AI neurons, AI synapses, AI regions, and AI systems respectively.

2) Most AI models focus on system implementation of different applications[33,34,57,58,59]. In this scenario, infinite numbers of functions in the brain would require infinite number of corresponding AI models, each likely with black-box architectures. This paper emphasizes AI techniques at the cellular level and proposes AI twins of the brain. Theoretically, just like the brain, infinite number of functions could arise from its white-box AI twins. AI twins could offer alternative solutions to brain disorders at the level of individual neurons, synapses, or small regions of the brain, provided that ethical concerns are addressed. In essence, AI theoretically has the capability of representing and substituting malfunctioning biological neurons and synapses in the brain (Theorem 1 and Theorem 2 in Appendix B). Priority for applications of alternative efficient solutions for brain diseases should ideally focus on those that specifically damage brain regions where carefully designed cell-level nanometer (and smaller) size of AI chips could help restore abnormal or mimic physiologic functions with bottom-up approaches from micrometer size of neurons and nanometer size of synapses[9] to pieces, regions, systems of the brain. Understanding the neurons level detailed structure information is critical for the future realization of such AI neuron replacement therapy, which depends on accurate measure of gain and loss of function at cell level. If brain structures could be scanned at the neural level, as demonstrated by recent developments and research efforts[11,60], this could potentially serve a similar role to today's MRI or CT scans for the identification of damaged or diseased neurons or areas, which could then be potentially replaced by AI components as described in this paper. AI could potentially detect or identify "gain of function" or "loss of function" signals at the neurons, synapses and proteins using smart materials and computational nodes.

## VII.  Appendices

### A. Function representations of biological neurons and biological synapses

Conventional mathematical models aim to explain the functional and operational mechanisms of biological neurons. It is not difficult to see that, even similar or single type of such models used for all neurons in the brain, the effort spent on combining billions of such mathematical models for billions of neurons in the brain would become endless and manpower intensive.

However, considering biological neurons and synapses are connected alternative in sequence, AI techniques could be used to represent biological neurons and synapses due to its universal learning capabilities[21,22,24].

Instead of focusing details of mathematical representations of spiking neurons and neuron dynamics molecular behaviors within neurons and synapses, both neurons and synapses have their essential models which are determined by their internal molecular behaviors. This paper considers that the function of each type of neuron and synapse are (piecewise) continuous and thus applies AI technique to model and represent the function of each type of neuron and synapse.

Definition 1 [61]
A function in one-dimensional space is piecewise continuous if it has only a finite number of discontinuities in any interval and its left and right limits are defined (not necessarily equal) at each discontinuity.

This definition can be extended to multi-dimensional spaces in which neurons and synapses operate.

Definition 2 [62]

A function $f$ in a multi-dimensional space / domain $\Omega$ is piecewise continuous if this space / domain can be divided into a finite number of disjoint regions $D_1, D_2, \cdots, D_k \subset \Omega$ such that $\Omega = \bigcup_{i=1}^{k} \overline{D_i}$ and for each region $D_i$ there is a function $g_i \in C(\overline{D_i})$ with $g_i = f$ on $D_i$.

This definition means that a function in a multi-dimensional space / domain is piecewise continuous if this space / domain can be divided into a finite number of disjoint regions with zero measure of boundaries and if the function is continuous within each of these regions but might have bounded (not infinite or oscillatory) discontinuities at the boundaries of regions.

Suppose that a biological neuron (indexed by $L$) has input signal $\mathbf{x}_L = (x_{L,1}, x_{L,2}, \cdots, x_{L,n})$ in its dendrite trees and the output signal of its presynaptic terminal is $\mathbf{y}_L = (y_{L,1}, y_{L,2}, \cdots, y_{L,m})$:

$$\mathbf{y}_L(\mathbf{x}) = A_L(\mathbf{x})$$

where $A_L(\mathbf{x})$ shows the input-output relationship functions between the input signals of neurons' dendrites and the output signals of axon terminals (Fig. 1). There may be hundreds of different types of potential action functions in 86 billion neurons of human brains. In essence, neurons are biological cells, and their action potential functions are (piecewise) continuous. And thus, the input-output relationship functions $A_L(\mathbf{x})$ between the input signals of dendrites and the output signals of axon terminals are naturally (piecewise) continuous as well due to biological reaction properties of neurons. After considering the probability and time delay in such relationship between membrane potential and channel state, the input-output relationship functions between the input signals of dendrites and the output signals of axon terminals could be explicitly written as:

$$\mathbf{y}_L(\mathbf{x}, P_L, t, \Delta_t) = A_L(\mathbf{x}, t)$$

This provides an alternative solution and representation of biological neurons which is different from the conventional spiking representation and neurons dynamics approaches[8].

Since the probability $P_L$, time $t$ and time delay $\Delta_t$ in such relationship, which are considered as the essential properties of neurons and part of the output values $\mathbf{y}_L$ of input-output relationship functions $A_L(\mathbf{x})$, may not affect the (piecewise) continuity property and all the discussions in this paper, the input-output relationship function of each single neuron could be simply written as $\mathbf{y}_L(\mathbf{x}) = A_L(\mathbf{x})$ without explicitly mentioning the probability $P_L$, time $t$ and time delay $\Delta_t$.

Like the action potentials of neurons, the neurotransmission relationship of $P$-th synapse between the presynaptic terminal at the end of an axon of a neuron and the postsynaptic terminal of the next connected neuron is $S_P(\mathbf{x})$, which is (piecewise) continuous as well[35]. This may be true to other types of connections (e.g., volume transmission, ehaptic coupling and electrical synapses, and other possible physical types of connections among neurons).

## B. AI representations of biological neurons and biological synapses

Hornik[21] rigorously proved a theorem showing that if the activation function is bounded nonconstant continuous, then any target continuous function could be approximated by single-hidden layer feedforward networks (SLFNs) over compact input sets. Without loss of generality, single-hidden layer feedforward networks (SLFN) are used in the basic theoretical analysis in this paper and all the analysis could be replaced with more efficient AI models whenever necessary.

*AI approximating single neuron and single synapse*

Let $L^2(X)$ be a space of functions $f$ on a compact subset $X$ in the $d$-dimensional Euclidean space $\mathbf{R}^d$ such that $|f|^2$ are integrable. The norm in $L^2(X)$ space will be denoted as $|\cdot|$, and the closeness between the network function $f_L$ and the target function $f$ is measured by the $L^2(X)$ distance: $|f_L - f| = \left[\int_X |f_L(\mathbf{x}) - f(\mathbf{x})|^2 \, d\mathbf{x}\right]^{1/2}$. A space $X$ being a compact subset $X$ of $\mathbf{R}^d$ means that such space $X$ is closed and bounded. A set being closed can be visualized as a set where all convergent sequences converge to a point in the set. The meaning of bounded is that there is an upper limit for the function.

Lemma 1 (Theorem 2 [21])
If $g$ is continuous, bounded and nonconstant, then $g(\mathbf{a} \cdot \mathbf{x} + b)$ is dense in $C(X)$ for all compact subset $X$ of $\mathbf{R}^d$, where $\mathbf{a} \cdot \mathbf{x} + b$ represents weighted summation of the input vector and $C(X)$ denotes the set of all continuous functions on compact subset $X$.

That is, given any small positive value $\delta > 0$, there exists $f_L$ with appropriate number $L$ of hidden nodes and properly adjusted hidden node parameters $(\mathbf{a}_i, b_i)$ such that $|f - f_L| = \left|f - \sum_{i=1}^L \beta_i \, g(\mathbf{a}_i \cdot \mathbf{x} + b_i)\right| < \delta$, where $\beta_i$ is the weight of the connection between the hidden node $i$ to the output node.

Hornik's theorem[21] is valid for any target piecewise continuous function $f$ and also any continuous, bounded and nonconstant activation function $g$. According to Lemma 1, given any input-output relationships of single neuron $A_L(\mathbf{x})$ and the neurotransmission relationships within their corresponding synapses $S_P(\mathbf{x})$, which are piecewise continuous, there exists an AI model (e.g., artificial feedforward neural networks) universally approximating them. In other words, the two fundamental computational elements (components) which form the human brain learning systems could be represented by single-hidden layer feedforward networks (SLFN) or a more efficient artificial neural network developed from and constructed by SLFNs with any small error.

Although without loss of generality, SLFNs are often employed in theoretical proofs, the realm of more efficient AI networks allows for combinations of basic SLFNs with diverse types of artificial neurons. Such artificial neurons could be convolutional neurons and many other types of neurons which are often used in different AI implementations nowadays. Since both fundamental input-output relationships of single neuron $A_L(\mathbf{x})$ and the neurotransmission relationships within their corresponding synapses $S_P(\mathbf{x})$ are (piecewise) continuous, they could be represented by AI with networks of artificial neurons. Thus, we have

Theorem 1
Given any input-output relationship of single neuron between its dendrite trees and the presynaptic terminal of its axon $A_L(\mathbf{x})$, there exists an artificial neural network, either SLFN or other types of efficient artificial neural networks, which could universally approximate $A_L(\mathbf{x})$ with any small error.

Theorem 2
Given any neurotransmission relationship within corresponding synapses $S_P(\mathbf{x})$, there exists an artificial neural network, either SLFN or other types of efficient artificial neural networks, which could universally approximate $S_P(\mathbf{x})$ with any small error.

## C. Theoretical proof of representation capabilities of bottom-up AI techniques

Neurons adhere to the all-or-none law that if a neuron responds it must respond completely. In terms of mathematical representation, such all-or-none smoothness law could be covered by the following definition:

Definition 3
A function $f$ is called all-or-none smoothness if given any value $c \in (\min f(\mathbf{x}), \max f(\mathbf{x}))$ and positive small value $\delta$, there exists zero or finite number of points $\mathbf{x}_1, \mathbf{x}_2, \cdots, \mathbf{x}_n$ such that:
1) $f(\mathbf{x}_i) = c$, and
2) for any small $|\Delta| < \delta$, there exist $\mathbf{x}_{i,1}$ and $\mathbf{x}_{i,2}$ in regions $(\mathbf{x}_i - \Delta, \mathbf{x}_i + \Delta)$ such that $f(\mathbf{x}_{i,1}) > c$ and $f(\mathbf{x}_{i,2}) < c$.

These finite points are called irregular points.

Theorem 3
Suppose that all the functions $f, f_1, \cdots, f_k$ are all-or-none smoothness piecewise continuous, then $f(f_1, f_2, \cdots, f_k)$ are all-or-none smoothness piecewise continuous.

*Proof*: Given any point $\mathbf{x}^*$ and any small value $\delta > 0$. As $f(\mathbf{x}_1, \mathbf{x}_2, \cdots, \mathbf{x}_k)$ is piecewise continuous, there exists $\Delta > 0$ such that $|f(\mathbf{x}) - f(\mathbf{x}^*)| < \delta$ when $|\mathbf{x} - \mathbf{x}^*| < \Delta$. As $f_1(\mathbf{x}), \cdots, f_k(\mathbf{x})$ are piecewise continuous, there exists small value $\Delta' > 0$ such that $|f_i(\mathbf{x}) - f_i(\mathbf{x}^*)| < \sqrt{\Delta/k}$ when $|\mathbf{x} - \mathbf{x}^*| < \Delta'$. Thus, $|f(f_1(\mathbf{x}), f_2(\mathbf{x}), \cdots, f_k(\mathbf{x})) - f(f_1(\mathbf{x}^*), f_2(\mathbf{x}^*), \cdots, f_k(\mathbf{x}^*))| < \delta$ when $|\mathbf{x} - \mathbf{x}^*| < \Delta'$.

Since $f, f_1, \cdots, f_k$ are all-or-none smoothness, apparently $f(f_1, f_2, \cdots, f_k)$ is all-or-none smoothness as above defined due to finite irregular points. This completes the proof.

Theorem 3 shows a general fact that when all-or-none smoothness piecewise continuous functions are combined using another all-or-none smoothness piecewise continuous function, the resultant composite function retains the all-or-none smoothness piecewise continuous property. This obviously includes the cases when biological neurons and synapses are sequentially connected alternatively one by one in order when Theorem 3 is applied recursively in such sequences.

Theorem 4
The human brain and any of its constituent regions constructed by sequentially linked neurons and synapses in the brain are all-or-none smoothness piecewise continuous.

*Proof*: Any regions of the brain are composed of two fundamental elements (components): neurons and synapses. All the input-output relationships of single neuron between its dendrite trees and the presynaptic terminal of its axon $A_L(\mathbf{x})$ and the relationships within their corresponding synapses $S_P(\mathbf{x})$ are all-or-none smoothness piecewise continuous. According to Theorem 3, such regions constructed by these two specific elements are all-or-none smoothness piecewise continuous as well. This completes the proof.

According to Theorem 4 in Appendix C, the human brain and any of its regions formed by neurons and synapses are all-or-none smoothness piecewise continuous, after each of its neurons and synapses is replaced and represented by appropriate AI components we have

Theorem 5
In theory, the human brain and any of its subsystems constructed by sequentially linked neurons and synapses could be represented and universally approximated by artificial neural networks with any small error.

### D. Essence of BP algorithm

Error backpropagate (BP) algorithm[49] is used to adjust parameters of artificial neural networks. BP algorithm is essentially an iterative gradient descent mathematical approach. BP algorithm consists of two steps: forward computation of signals from the input layer to the output layer, layer by layer, and backward computation of errors from the output layer to the first layer through hidden layers, layer by layer.

*Forward computation*

$$in_i = \sum_j W_{j,i} a_j$$

where $a_j$ is the output of node $j$, $W_{j,i}$ is the weight of the connection from node $j$ in the preceding layer to node $i$ in the next layer, and $in_i$ is the input of node $i$.

*Backward computation*

Until performance is satisfactory, weights of connections in artificial neural networks are adjusted through examples iteratively over multiple epochs:

Output nodes:
$$W_{j,i} \leftarrow W_{j,i} + \alpha \times a_j \times \Delta_i$$
$$\Delta_i = Err_i \times g'(in_i)$$

where $g$ is the activation function of computational nodes, $Err_i$ is the error between the expected target and the true output of the output node $i$, $\Delta_i$ is the error resulted from the input signal of node $i$.

Hidden nodes

$$W_{k,j} \leftarrow W_{k,j} + \alpha \times a_k \times \Delta_j$$

$$\Delta_j = g'(in_j) \sum_i W_{j,i} \Delta_i$$

$$Err_j = \sum_i W_{j,i} \Delta_i$$

$Err_j$ is the sum of all the errors backwarded from all the nodes in next layer directly connected to node $j$.

Seen from the forward computation and backward computation steps, in order to adjust the parameters of networks BP algorithm requests that connections among artificial neural networks need to be bidirectional for concurrent computations of signal forward and error backward (Fig. 7).

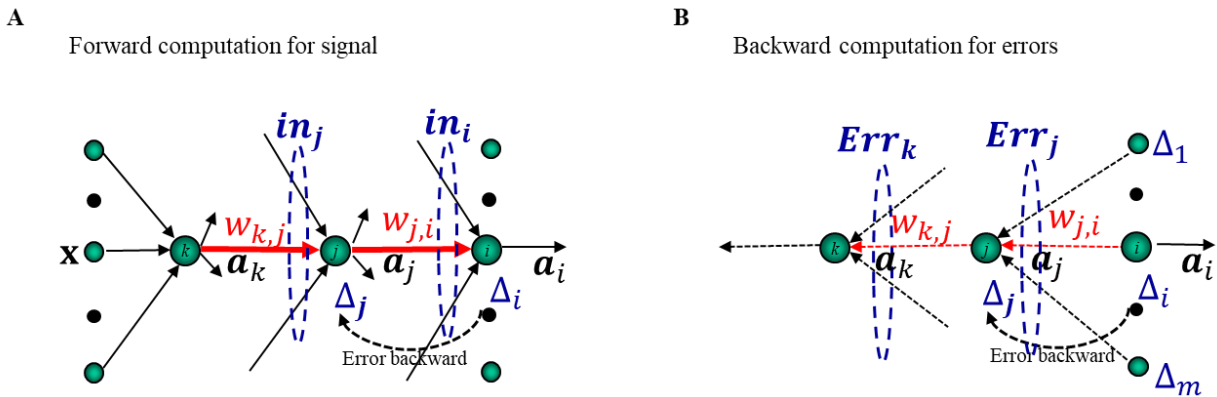

**Fig. 7. Error backpropagation (BP) algorithm is not used to tune parameters of neurons and synapses in the brain.** (**A**) Signals are transferred to the artificial neurons in next hidden layers through connections (wires). (**B**) Errors are back propagated to artificial neurons in earlier layers through same connections (same wires). Connections between adjacent artificial neurons concurrently transfer signals and propagate errors in two ways when BP algorithm is used, however, transmissions in neurons and synapses in the brain are unidirectional.

### E. Potential appropriate AI governance approach

In general, AI grows exponentially. AI would inevitably have serious impact on social order and ethics[63]. AI would surpass human brain capabilities in the end if there were no proper restrictions and governances in the aspects of either legal and ethical regulations or technologies (Fig. 8 and Fig. 9). From the perspective of technologies, it is not straightforward to have restrictions on those key properties which are going to become more and more invisible in most cases, e.g., algorithms (and numbers of their parameters), data sources, smart materials, AI agents, and knowledge exchanges for AI. AI algorithms may self-generate new algorithms with varying architectures and models, data are complicated and invisible in most cases, smart materials are becoming popular in AI implementations, knowledge exchange among AI models and AI agents are not easily controllable. Out of the above mentioned six key properties, computing power may be readily manageable, measurable and explainable to humans in two aspects:
1) The upper bound of computing capacities.

2) The upper bound of number of neurons of AI models in which artificial neural networks play significant roles or equivalent scales in other types of AI models (in terms of number of neurons).

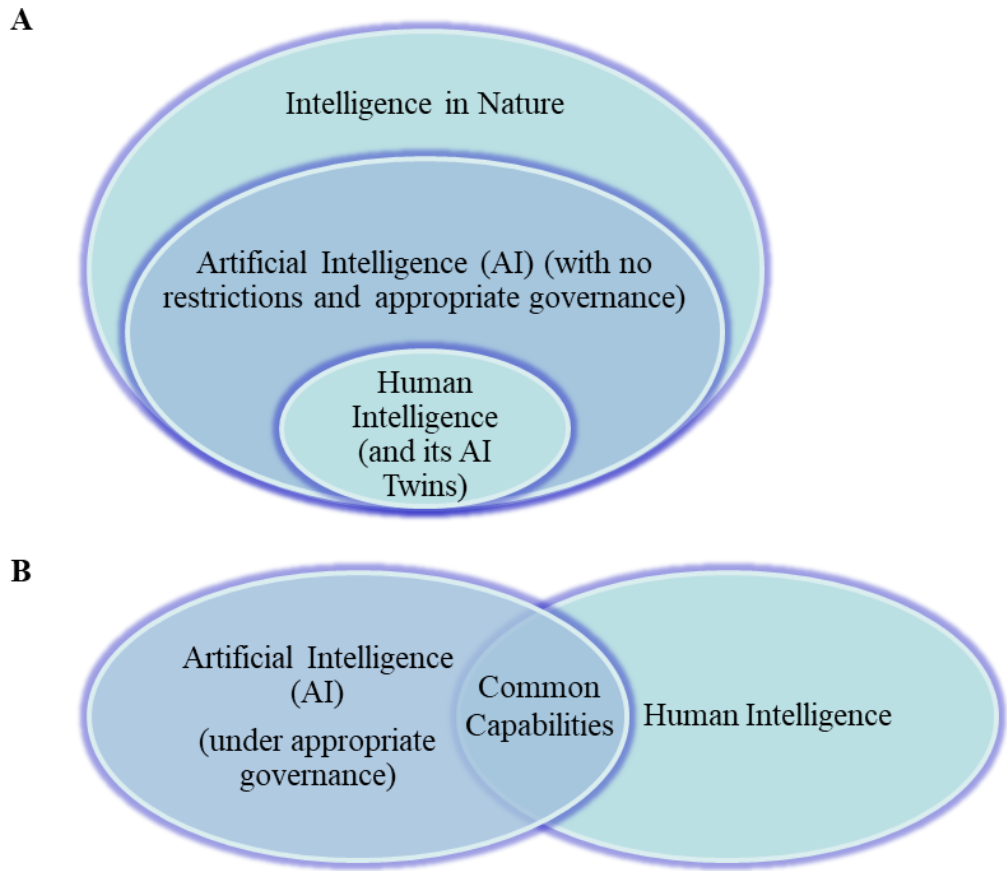

**Fig. 8. Artificial intelligence (AI) versus human intelligence.** (**A**) Artificial intelligence would surpass human intelligence and human intelligence would become a small subset of growing Intelligence if AI develops without restrictions and appropriate governance. AI would self-grow and develop further and Intelligence moves forward beyond human's imagination and capability. (**B**) AI and human intelligence share some common capabilities while having different strengths if AI is under appropriate governance.

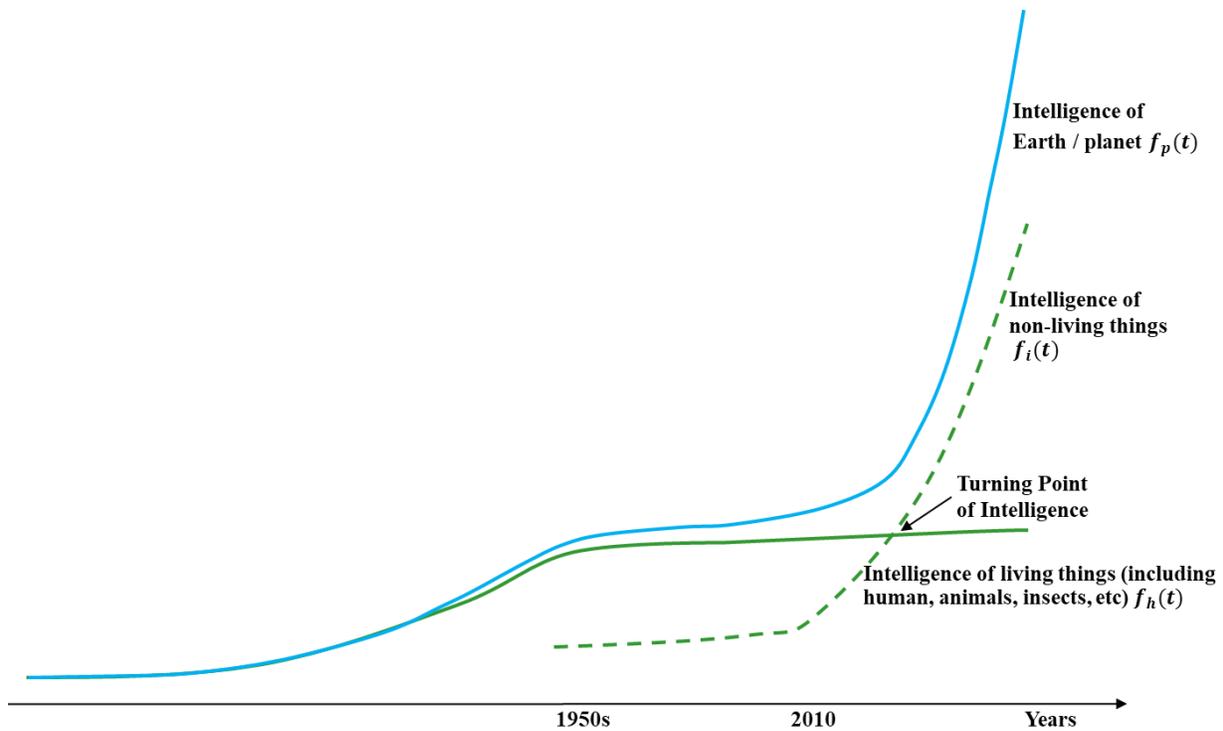

**Fig. 9. Exponential growth of AI.** Turning points of Intelligence if AI without restrictions and appropriate governance.